\documentclass[11pt,a4paper]{article}
\usepackage[hyperref]{acl2017}
\usepackage{times}
\usepackage{latexsym}
\usepackage{times}
\usepackage{stfloats}
\usepackage{url}
\usepackage{latexsym}
\usepackage{times}
\usepackage{float}
\usepackage{url}
\usepackage{graphicx}
\usepackage{epstopdf}
\usepackage{amsfonts}
\usepackage{amssymb}
\usepackage{fixltx2e}
\usepackage{amsmath}
\usepackage{verbatim}
\usepackage{multirow}
\usepackage{subfigure}
\usepackage{pifont}% http://ctan.org/pkg/pifont
\usepackage{amssymb}% http://ctan.org/pkg/amssymb
\usepackage{array}
\usepackage{fp}
\usepackage{makecell}
\usepackage{flushend}
\usepackage{cases}
\usepackage{color}
\usepackage{bbm}
\newcommand{\thickhline}{\noalign{\hrule height 1pt}}
\def\aclpaperid{37} 
\newcommand{\xmark}{\ding{55}}%

\aclfinalcopy

\title{Sequential Matching Network: A New Architecture for Multi-turn Response Selection in Retrieval-Based Chatbots}

\author{
	Yu Wu$^\dag$, Wei Wu$^\ddag$,	Chen Xing$^\diamondsuit$, Zhoujun Li$^\dag$\thanks{~~~Corresponding Author}~, Ming Zhou$^\ddag$~~~~\\
	$^\dag$State Key Lab of Software Development Environment, Beihang University, Beijing, China\\
	$^ \diamondsuit$College of Computer and Control Engineering, Nankai University, Tianjin, China\\
	$^\ddag$~~~~Microsoft Research, Beijing, China\\
	\{wuyu,lizj\}@buaa.edu.cn \{wuwei,v-chxing,mingzhou\}@microsoft.com 
}
\date{}

\begin{document}
	\maketitle
	\begin{abstract}
		We study response selection for multi-turn conversation in retrieval-based chatbots. Existing work either concatenates utterances in context or matches a response with a highly abstract context vector finally, which may lose relationships among utterances or important contextual information. We propose a sequential matching network (SMN) to address both problems. SMN first matches a response with each utterance in the context on multiple levels of granularity, and distills important matching information from each pair as a vector with convolution and pooling operations. The vectors are then accumulated in a chronological order  through a recurrent neural network (RNN) which models relationships among utterances. The final matching score is calculated with the hidden states of the RNN. An empirical study on two public data sets shows that SMN can significantly outperform state-of-the-art methods for response selection in multi-turn conversation.
	\end{abstract}
	
	\section{Introduction}
	%Human-computer conversation is an important task in NLP and AI.  
	Conversational agents include task-oriented dialog systems and non-task-oriented chatbots. Dialog systems focus on helping people complete specific tasks in vertical domains \cite{young2010hidden}, while chatbots aim to naturally and meaningfully converse with humans on open domain topics \cite{ritter2011data}. Existing work on building chatbots includes generation -based methods and retrieval-based methods. Retrieval based chatbots enjoy the advantage of informative and fluent responses, because they select a proper response for the current conversation from a repository with response selection algorithms.  While most existing work on retrieval-based chatbots studies response selection for single-turn conversation \cite{wang2013dataset} which only considers the last input message, we consider the problem in a multi-turn scenario. In a chatbot, multi-turn response selection takes a message and utterances in its previous turns as input and selects a response that is natural and relevant to the whole context.
	
	%In this paper, we study multi-turn response selection of retrieval based chatbot which takes all of conversation history as input rather than the last turn of conversation. 
	\begin{table}[!t]
		
		\label{example1} \small 
		\centering	
			\scalebox{0.8}{
		\begin{tabular}{l|l}
			\hline
		   &\textbf{Context} \\ \thickhline
			utterance 1& \emph{Human}: How are you doing? \\ \hline
		utterance 2&	\emph{ChatBot}: I am going to \textbf{hold a drum class} in Shanghai. \\ &Anyone wants to join? The location is near Lujiazui.\\ \hline
			%Response 1 : Yes, I did. \\ \hline
		utterance 3&	\emph{Human}: Interesting! Do you have coaches who\\& can help me practice \textbf{drum}? \\ \hline
		utterance 4&	\emph{ChatBot}: Of course. \\ \hline
		utterance 5&	\emph{Human}: Can I have a free first lesson?\\  \thickhline
		&	\textbf{Response Candidates} \\ \hline
		response 1&	 Sure. Have you ever played drum before? \checkmark  \\ \hline
		response 2&	 What lessons do you want? \xmark \\ \hline
			
			%Response: If you truly want to learn drum,\\ you can directly contact me by phone (number). \\
		\end{tabular}
	}
		\caption{An example of multi-turn conversation}	\vspace{-5mm}
	\end{table}
	The key to response selection lies in input-response matching. Different from single-turn conversation, multi-turn conversation requires matching between a response and a conversation context %\footnote{In this paper, both ``session'' and ``context'' refer to an input message and utterances in its previous turns.} 
	in which one needs to consider not only the matching between the response and the input message but also  matching between responses and utterances in previous turns.  The challenges of the task include (1) how to identify important information (words, phrases, and sentences) in context, which is crucial to selecting a proper response and leveraging relevant information in matching; and (2) how to model relationships among the utterances in the context. %How much important information an utterance contains also indicates its importance to response selection. 
	Table \ref{example1} illustrates the challenges with an example.  First, ``hold a drum class'' and ``drum'' in context are very important. %to response selection for the session. 
	Without them, one may find responses relevant to the message (i.e., the fifth utterance of the context) but nonsense in the context (e.g., ``what lessons do you want?'').  %On the other hand, although ``Shanghai'' and ``Lujiazui'' are keywords in their utterances, they are useless and even noise to response selection. It is crucial yet non-trivial to extract the important information from the context and leverage them in matching while circumvent the noise. 
	Second, the message highly depends on the second utterance in the context, and the order of the utterances matters in response selection: exchanging the third utterance and the fifth utterance may lead to different responses. %Third, because Context 1 contains much important information, it is more important than others in response selection. 
	Existing work, however, either ignores relationships among utterances when concatenating them together \cite{lowe2015ubuntu}, or loses important information in context in the process of converting the whole context to a vector without enough supervision from responses (e.g., by a hierarchical RNN \cite{zhou2016multi}).     
	
	We propose a sequential matching network (SMN), a new context based matching model that can tackle both challenges in an end-to-end way. The reason that existing models lose important information in the context is that they first represent the whole context as a vector and then match the context vector with a response vector. Thus, responses in these models connect with the context until the final step in matching. To avoid information loss, SMN matches a response with each utterance in the context at the beginning and encodes important information in each pair into a matching vector. The matching vectors are then accumulated in the utterances' temporal order to model their relationships. The final matching degree is computed with the accumulation of the matching vectors. Specifically, for each utterance-response pair, the model constructs a word-word similarity matrix and a sequence-sequence similarity matrix by the word embeddings and the hidden states of a recurrent neural network with gated recurrent units (GRU) \cite{chung2014empirical} respectively. The two matrices capture important matching information in the pair on a word level and a segment (word subsequence) level respectively, and the information is distilled and fused as a matching vector through an alternation of convolution and pooling operations on the matrices. By this means, important information from multiple levels of granularity in context is recognized under sufficient supervision from the response and carried into matching with minimal loss. The matching vectors are then uploaded to another GRU to form a matching score for the context and the response. The GRU accumulates the pair matching in its hidden states in the chronological order of the utterances in context. It models relationships and dependencies among the utterances in a matching fashion and has the utterance order supervise the accumulation of pair matching. %The gate mechanism of the GRU helps select important pairs and filter out noise. 
	 The matching degree of the context and the response is computed by a logit model with the hidden states of the GRU. SMN extends the powerful ``2D'' matching paradigm in text pair matching for single-turn conversation to context based matching for multi-turn conversation, and enjoys the advantage of both important information in utterance-response pairs and relationships among utterances being sufficiently preserved and leveraged in matching.

	We test our model on the Ubuntu dialogue corpus \cite{lowe2015ubuntu} which is a large scale publicly available English data set for research in multi-turn conversation. The results show that our model can significantly outperform state-of-the-art methods, and  improvement to the best baseline model on R$_{10}$@1 is over $6$\%. In addition to the Ubuntu corpus, we create a human-labeled Chinese data set, namely the Douban Conversation Corpus, and test our model on it. In contrast to the Ubuntu corpus in which data is collected from a specific domain and negative candidates are randomly sampled, conversations in this data come from the open domain, and response candidates in this data set are collected from a retrieval engine and labeled by three human judges. On this data, our model improves the best baseline model by $3$\%  on R$_{10}$@1 and $4$\% on P@1. As far as we know, Douban Conversation Corpus is the first human-labeled data set for multi-turn response selection and could be a good complement to the Ubuntu corpus. We have released Douban Conversation Corups and our source code at \url{https://github.com/MarkWuNLP/MultiTurnResponseSelection} %\url{https://github.com/MarkWuNLP/MultiTurnResponseSelection}.
	
	% One problem with the Ubuntu data is that negative examples in test are randomly sampled which might oversimplify the multi-turn problem in a real retrieval based chatbot. To further verify the efficacy of the proposed model in a real situation, we simulate the procedure of a retrieval based chatbot and create a large scale Chinese test set. Instead of negative sampling, labels in the data are generated by $3$ human judges. On this data, our model improves the best baseline model over $4$\% on P@1 (equivalent to R$_{10}$@1).  We published the data at \url{https://github.com/MarkWuNLP/MultiTurnResponseSelection}.

	Our contributions in this paper are three-folds: (1) the proposal of a new context based matching model for multi-turn response selection in retrieval-based chatbots; (2) the publication of a large human-labeled data set to research communities; (3) empirical verification of the effectiveness of the model on public data sets.

	\section{Related Work}
	%Early work \cite{weizenbaum1966eliza} on chatbots exploits hand crafted templates to generate responses, which requires huge human effort and is not scalable. 
	%Recently, building a chatbot with data driven approaches \cite{ritter2011data,higashinaka2014towards} has drawn a lot of attention. Existing work along this line includes retrieval based methods and generation based methods. The former selects a proper response from an index based on matching between the response and an input message with or without context \cite{hu2014convolutional,ji2014information,wang2015syntax,DBLP:conf/sigir/YanSW16,DBLP:journals/corr/WuWLZ16,zhou2016multi}, while the latter employs statistical machine translation techniques \cite{ritter2011data} or the sequence to sequence framework \cite{DBLP:conf/acl/ShangLL15,serban2015building,vinyals2015neural,li2015diversity,li2016persona,xing2016topic,serban2016multiresolution} to generate responses. Our work belongs to retrieval based methods, and we study response selection with context information.
	
	Recently, building a chatbot with data driven approaches \cite{ritter2011data,ji2014information} has drawn significant attention. Existing work along this line includes retrieval-based methods \cite{hu2014convolutional,ji2014information,wang2015syntax,DBLP:conf/sigir/YanSW16,DBLP:journals/corr/WuWLZ16,zhou2016multi,wu2016ranking} and generation-based methods \cite{DBLP:conf/acl/ShangLL15,serban2015building,vinyals2015neural,li2015diversity,li2016persona,xing2016topic,serban2016multiresolution}. Our work is a retrieval-based method, in which we study context-based response selection. 
	
	Early studies of retrieval-based chatbots focus on response selection for single-turn conversation \cite{wang2013dataset,ji2014information,wang2015syntax,DBLP:journals/corr/WuWLZ16}.
	Recently, researchers have begun to pay attention to multi-turn conversation. For example, Lowe et al. \shortcite{lowe2015ubuntu} match a response with the literal concatenation of context utterances. Yan et al. \shortcite{DBLP:conf/sigir/YanSW16} concatenate context utterances with the input message as reformulated queries and perform matching with a deep neural network architecture. Zhou et al. \shortcite{zhou2016multi} improve multi-turn response selection with a multi-view model including an utterance view and a word view. % The stark difference between our model and the existing models is that our model matches a response with each utterance at the very first and matching information instead of sentences is accumulated in a temporal manner through a GRU. 
	Our model is different in that it matches a response with each utterance at first and accumulates matching information instead of sentences by a GRU, thus useful information for matching can be sufficiently retained.

	\section{Sequential Matching Network}
	%We elaborate our matching approach for multi-turn conversation in retrieval based chatbots.
	\subsection{Problem Formalization}\label{probform}
	Suppose that we have a data set $\mathcal {D} = \{(y_i,s_i,r_i)\}_{i=1}^N$, where $s_i=\{u_{i,1}, \ldots, u_{i,n_i}\}$ represents a conversation context with $\{u_{i,k}\}_{k=1}^{n_i}$ as utterances.  $r_i$ is a response candidate and $y_i\in \{0,1\}$ denotes a label. $y_i=1$ means $r_i$ is a proper response for $s_i$, otherwise $y_i=0$.  Our goal is to learn a matching model $g(\cdot,\cdot)$ with $\mathcal{D}$. For any context-response pair $(s,r)$, $g(s,r)$ measures the matching degree between $s$ and $r$. 
	
	\subsection{Model Overview}
	\begin{figure*}[t]	
		\begin{center}
			\includegraphics[width=14cm,height=5.5cm]{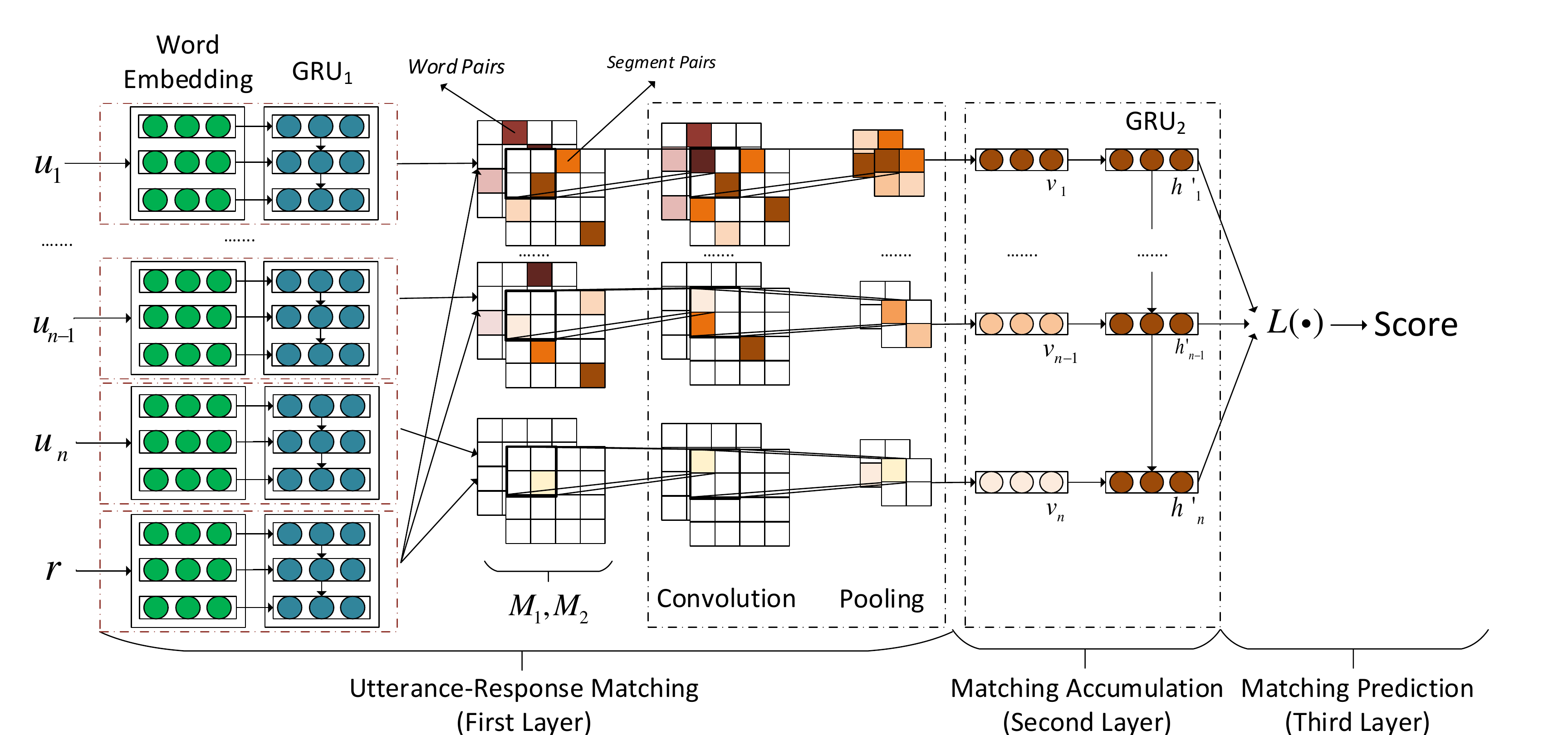}	
		\end{center}
		\vspace{-3mm}
		\caption{Architecture of SMN}\label{fig:arch}\vspace{-3mm}
	\end{figure*}
	We propose a sequential matching network (SMN) to model $g(\cdot,\cdot)$. Figure \ref{fig:arch} gives the architecture. SMN first decomposes context-response matching into several utterance-response pair matching and then all pairs matching are accumulated as a context based matching through a recurrent neural network. SMN consists of three layers. The first layer matches a response candidate with each utterance in the context on a word level and a segment level, and important matching information from the two levels is distilled by convolution, pooling and encoded in a matching vector. %An utterance-response pair is transformed to a word-word similarity matrix and a sequence-sequence similarity matrix,  
	The matching vectors are then fed into the second layer where they are accumulated in the hidden states of a recurrent neural network with GRU following the chronological order of the utterances in the context. The third layer calculates the final matching score with the hidden states of the second layer.
	
	SMN enjoys several advantages over existing models. First, a response candidate can match each utterance in the context at the very beginning, thus matching information in every utterance-response pair can be sufficiently extracted and carried to the final matching score with minimal loss. Second, information extraction from each utterance is conducted on different levels of granularity and under sufficient supervision from the response, thus semantic structures that are useful for response selection in each utterance can be well identified and extracted. Third, matching and utterance relationships are coupled rather than separately modeled, thus utterance relationships (e.g., order), as a kind of knowledge, can supervise the formation of the matching score.  
	
	By taking utterance relationships into account, SMN extends the ``2D'' matching that has proven effective in text pair matching for single-turn response selection to sequential ``2D'' matching for context based matching in response selection for multi-turn conversation. In the following sections, we will describe details of the three layers. 
	
	\subsection{Utterance-Response Matching}\label{multi-channel}
	Given an utterance $u$ in a context $s$ and a response candidate $r$, the model looks up an embedding table and represents $u$ and $r$ as $\mathbf{U} = \left[e_{u,1},\ldots,e_{u,n_u}\right]$ and $\mathbf{R} = \left[e_{r,1},\ldots,e_{r,n_r}\right]$ respectively, where $e_{u,i}, e_{r,i} \in \mathbb{R} ^ d$ are the embeddings of the $i$-th word of $u$ and $r$ respectively. 
	$\mathbf{U}$ $ \in \mathbb{R}^{d \times n_u}$ and $\mathbf{R}$  $ \in \mathbb{R}^{d \times n_r}$ are then used to construct a word-word similarity matrix $\mathbf{M}_1$  $ \in \mathbb{R}^{n_u \times n_r}$ and a sequence-sequence similarity matrix $\mathbf{M}_2$ $ \in \mathbb{R}^{n_u \times n_r}$  which are two input channels of a convolutional neural network (CNN). The CNN distills important matching information from the matrices and encodes the information into a matching vector $v$. 
	
	Specifically, $\forall i,j$, the $(i,j)$-th element of $\mathbf{M}_1$ is defined by
	\begin{equation}\label{M1Element}\small
	e_{1,i,j} = e_{u,i}^{\top} \cdot e_{r,j}. 
	\end{equation}
	$\mathbf{M}_1$ models the matching between $u$ and $r$ on a word level. 
	
	To construct $\mathbf{M}_2$, we first employ a GRU to transform $\mathbf{U}$ and $\mathbf{R}$ to hidden vectors. Suppose that $\mathbf{H}_u = \left[h_{u,1},\ldots, h_{u,n_u}\right]$ are the hidden vectors of $\mathbf{U}$, then $\forall i$, $h_{u,i}\in \mathbb{R}^m$ is defined by 
	
	\small	 
	\begin{eqnarray}\label{gru}
	\small
	&& z_i = \sigma(\mathbf{W_z} e_{u,i} + \mathbf{U_z} {h}_{u,i-1}) \nonumber \\
	&& r_i = \sigma(\mathbf{W_r} e_{u,i} + \mathbf{U_r} {h}_{u,i-1}) \nonumber \\
	&&　\widetilde{h}_{u,i} = tanh(\mathbf{W_h} e_{u,i} + \mathbf{U_h} (r_i \odot {h}_{u,i-1}))\nonumber　\\
	&& h_{u,i} = z_i \odot \widetilde{h}_{u,i} + (1-z_i) \odot {h}_{u,i-1},
	\end{eqnarray}
	\normalsize	 	 
	where $h_{u,0}=0$, $z_i$ and $r_i$ are an update gate and a reset gate respectively, $\sigma(\cdot)$ is a sigmoid function, and $\mathbf{W_z}$, $\mathbf{W_h}$, $\mathbf{W_r}$, $\mathbf{U_z}$, $\mathbf{U_r}$,$ \mathbf{U_h}$ are parameters. Similarly, we have $\mathbf{H}_r = \left[h_{r,1},\ldots, h_{r,n_r}\right]$ as the hidden vectors of $\mathbf{R}$. 
	Then, $\forall i,j$,  the $(i,j)$-th element of $\mathbf{M}_2$ is defined by
	\begin{equation}\label{M2Element}
	\small
	e_{2,i,j} = h_{u,i}^{\top} \mathbf{A}  h_{r,j},
	\end{equation}	 
	where $\mathbf{A} \in \mathbb{R}^{m \times m}$ is a linear transformation. $\forall i$, GRU models the sequential relationship and the dependency among words up to position $i$ and encodes the text segment until the $i$-th word to a hidden vector. Therefore, $\mathbf{M}_2$ models the matching between $u$ and $r$ on a segment level.    
	
	$\mathbf{M}_1$ and $\mathbf{M}_2$ are then processed by a CNN to form $v$. $\forall f=1,2$, CNN regards $\mathbf{M}_f$ as an input channel, and alternates convolution and max-pooling operations. Suppose that $z^{(l,f)} = \left[ z^{(l,f)}_{i,j}  \right]_{I^{(l,f)} \times J^{(l,f)}}$ denotes the output of feature maps of type-$f$ on layer-$l$, where $z^{(0,f)}= \mathbf{M}_f$, $\forall f = 1,2$. On the convolution layer, we employ a 2D convolution operation with a window size ${r_w^{(l,f)} \times r_h^{(l,f)}}$, and define $z_{i,j}^{(l,f)}$ as
	\begin{equation}\label{Conv}\small
	z_{i,j}^{(l,f)} = \sigma (\sum_{f'=0}^{F_{l-1}} \sum_{s=0}^{r_w^{(l,f)}} \sum_{t=0}^{r_h^{(l,f)}} \mathbf{W}_{s,t} ^ {(l,f)} \cdot z_{i+s,j+t} ^{(l-1,f')} + \mathbf{b}^{l,k} ), 
	\end{equation}
	where $\sigma(\cdot)$ is a ReLU, $\mathbf{W} ^ {(l,f)} \in \mathbb{R}^{r_w^{(l,f)} \times r_h^{(l,f)}} $ and $\mathbf{b}^{l,k}$ are parameters, and $F_{l-1}$ is the number of feature maps on the $(l-1)$-th layer. A max pooling operation follows a convolution operation and can be formulated as 
	\begin{equation}\label{Pool}\small
	z_{i,j}^{(l,f)} = \max_{p_w^{(l,f)} >  s \geq 0} \max_{p_h^{(l,f)} > t \geq 0} z_{i+s,j+t} , 
	\end{equation}
	where $p_w^{(l,f)}$ and $p_h^{(l,f)}$ are the width and the height of the 2D pooling respectively. The output of the final feature maps are concatenated and mapped to a low dimensional space with a linear transformation as the matching vector $v \in \mathbb{R}^q$.
	
	According to Equation (\ref{M1Element}), (\ref{M2Element}), (\ref{Conv}), and (\ref{Pool}), we can see that by learning word embedding and parameters of GRU from training data, words or segments in an utterance that are useful for recognizing the appropriateness of a response may have high similarity with some words or segments in the response and result in high value areas in the similarity matrices. These areas will be transformed and selected by convolution and pooling operations and carry important information in the utterance to the matching vector. This is how our model identifies important information in context and leverage it in matching under the supervision of the response. We consider multiple channels because we want to capture important matching information on multiple levels of granularity of text.    
	
	\subsection{Matching Accumulation}\label{match_acum}
	Suppose that $\left[v_1,\ldots,v_n\right]$ is the output of the first layer (corresponding to $n$ pairs), at the second layer, a GRU takes $\left[v_1,\ldots,v_n\right]$ as an input and encodes the matching sequence into its hidden states $H_m = \left[h'_1,\ldots, h'_n\right] \in \mathbb{R}^{q \times n}$ with a detailed parameterization similar to Equation (\ref{gru}). This layer has two functions: (1) it models  the dependency and the temporal relationship of utterances in the context; (2) it leverages the temporal relationship to supervise the accumulation of the pair matching as a context based matching.  Moreover, from Equation (\ref{gru}), we can see that the reset gate (i.e., $r_i$) and the update gate (i.e., $z_i$) control how much information from the previous hidden state and the current input flows to the current hidden state, thus important matching vectors (corresponding to important utterances) can be accumulated while noise in the vectors can be filtered out.

	\subsection{Matching Prediction and Learning}
	With $\left[h'_1,\ldots, h'_n\right]$, we define $g(s,r)$ as
	\begin{equation}\small
	g(s,r) = softmax (\mathbf{W_2} L[h'_1,\ldots, h'_n] + \mathbf{b_2}),
	\end{equation}	where $\mathbf{W_2}$ and $\mathbf{b_2}$ are parameters.
	We consider three parameterizations for $L[h'_1,\ldots, h'_n]$: (1) only the last hidden state is used. Then $L[h'_1,\ldots, h'_n]=h'_n$. (2) the hidden states are linearly combined. Then, $L[h'_1,\ldots, h'_n]=\sum_{i=1}^{n} w_i h'_i$, where $w_i \in \mathbb{R}$. (3) we follow \cite{yang2016hierarchical} and employ an attention mechanism to combine the hidden states. Then, $L[h'_1,\ldots, h'_n]$ is defined as

\begin{small}
	\begin{eqnarray}
		&& t_i = tanh(\mathbf{W_{1,1}} h_{u_i,n_u} + \mathbf{W_{1,2}} h'_i + \mathbf{b_1}),\nonumber \\
		&& \alpha_i = \frac{exp(t_i^{\top} t_s)}{\sum_i(exp(t_i^{\top} t_s))},\nonumber \\
		&&　L[h'_1,\ldots, h'_n]= \sum_{i=1}^{n}{\alpha_i h'_i},
	\end{eqnarray}
	\end{small}where $\mathbf{W_{1,1}} \in \mathbb{R}^{q \times m}, \mathbf{W_{1,2}} \in \mathbb{R}^{q \times q}$ and $\mathbf{b_1} \in \mathbb{R}^q$ are parameters. $h'_i$ and $h_{u_i,n_u}$ are the $i$-th matching vector and the final hidden state of the $i$-th utterance respectively. %We believe both matching vector and utterance benefit to recognize important utterances for matching. 
	$t_s \in \mathbb{R}^q$ is a virtual context vector which is randomly initialized and jointly learned in training.
	
	Both (2) and (3) aim to learn weights for $\{h'_1,\ldots,h'_n\}$ from training data and highlight the effect of important matching vectors in the final matching. The difference is that weights in (2) are static, because the weights are totally determined by the positions of utterances, while weights in (3) are dynamically computed by the matching vectors and utterance vectors. We denote our model with the three parameterizations of $L[h'_1,\ldots, h'_n]$ as SMN$_{last}$, SMN$_{static}$, and SMN$_{dynamic}$, and empirically compare them in experiments.

	We learn $g(\cdot, \cdot)$ by minimizing cross entropy with $\mathcal{D}$. Let $\Theta$ denote the parameters of SMN, then the objective function $\mathcal{L}(\mathcal{D},\Theta)$ of learning can be formulated as  
	\begin{equation}\label{obj}\small
	- \sum_{i=1}^{N} \left[y_i log(g(s_i,r_i)) + (1-y_i)log(1-g(s_i,r_i))\right].
	\end{equation}
	%where $N$ in the number of instances in $\mathcal{D}$. %We optimize the objective function using back-propagation and the parameters are updated by stochastic gradient descent with Adam algorithm \cite{kingma2014adam}. %We share our codes for training and test at \url{??}.

	\section{Response Candidate Retrieval} \label{candidate_retrieval}
	In practice, a retrieval-based chatbot, to apply the matching approach to the response selection, one needs to retrieve a number of response candidates from an index beforehand. While candidate retrieval is not the focus of the paper, it is an important step in a real system. In this work, we exploit a heuristic method to obtain response candidates from the index. Given a message $u_n$ with $\{u_1,\ldots,u_{n-1}\}$ utterances in its previous turns, we extract the top $5$ keywords from $\{u_1,\ldots,u_{n-1}\}$ based on their tf-idf  scores\footnote{Tf is word frequency in the context, while idf is calculated using the entire index.} and expand $u_n$ with the keywords. Then we send the expanded message to the index and retrieve response candidates using the inline retrieval algorithm of the index. Finally, we use $g(s,r)$ to re-rank the candidates and return the top one as a response to the context.  
	
	\section{Experiments}
	We tested our model on a publicly available English data set and a Chinese data set published with this paper.
	
	\subsection{Ubuntu Corpus}
	The English data set is the Ubuntu Corpus \cite{lowe2015ubuntu} which contains multi-turn dialogues collected from chat logs of the Ubuntu Forum. The data set consists of $1$ million context-response pairs for training, $0.5$ million pairs for validation, and $0.5$ million pairs for testing. Positive responses are true responses from humans, and negative ones are randomly sampled. The ratio of the positive and the negative is 1:1 in training, and 1:9 in validation and testing. We used the copy shared by Xu et al. \shortcite{xu2016incorporating} \footnote{\url{https://www.dropbox.com/s/2fdn26rj6h9bpvl/ubuntu data.zip?dl=0}} 
	in which numbers, urls, and paths are replaced by special placeholders. We followed \cite{lowe2015ubuntu} and employed recall at position $k$ in $n$ candidates ($R_n@k$) as evaluation metrics.
	\subsection{Douban Conversation Corpus}
	%	One problem with the Ubuntu data is that negative examples in test is much easier to identify than those in a real chatbot, because they are randomly sampled and most of them are far from the semantics of the context. A better test set that can simulate the real scenario of a retrieval based chatbot must have responses generated following the procedure of information retrieval and labels  annotated by humans. As far as we know, however, there are no such data sets publicly available. To test our model in a setting closer to the real case and facilitate the research of multi-turn response selection, we created a new data set and publish it to research communities with the paper.  
	The Ubuntu Corpus is a domain specific data set, and response candidates are obtained from negative sampling without human judgment. To further verify the efficacy of our model, we created a new data set with open domain conversations, called the Douban Conversation Corpus. Response candidates in the test set of the Douban Conversation Corpus are collected following the procedure of a retrieval-based chatbot and are labeled by human judges. It simulates the real scenario of a retrieval-based chatbot. We publish it to research communities to facilitate the research of multi-turn response selection.
	
	%	Ideally, we should use human-machine chatlog, however, chatlog often has privacy issues. Therefore we use human-human conversation as an alternation.  
	Specifically, we crawled $1.1$ million dyadic dialogues (conversation between two persons) longer than $2$ turns from Douban group\footnote{\url{https://www.douban.com/group}} which is a popular social networking service in China. We randomly sampled $0.5$ million dialogues for creating a training set, $25$ thousand dialouges for creating a validation set, and $1,000$ dialogues for creating a test set, and made sure that there is no overlap between the three sets. For each dialogue in training and validation,  we took the last turn as a positive response for the previous turns as a context and randomly sampled another response from the $1.1$ million data as a negative response. There are $1$ million context-response pairs in the training set and $50$ thousand pairs in the validation set. 
	
	To create the test set,  we  first crawled $15$ million post-reply pairs from Sina Weibo\footnote{\url{http://weibo.com/}} which is the largest microblogging service in China and indexed the pairs with Lucene\footnote{\url{https://lucenenet.apache.org/}}. We took the last turn of each Douban dyadic dialogue in the test set as a message, retrieved $10$ response candidates from the index following the method in Section \ref{candidate_retrieval}, and finally formed a test set with $10,000$ context-response pairs. We recruited three labelers to judge 
	if a candidate is a proper response to the context. A proper response means the response can naturally reply to the message given the whole context. Each pair received three labels and the majority of the labels were taken as the final decision. Table \ref{dataset} gives the statistics of the three sets. Note that the Fleiss' kappa \cite{fleiss1971measuring} of the labeling is $0.41$, which indicates that the three labelers reached a relatively high agreement.  
	
	Besides $R_n@k$s, we also followed the convention of information retrieval and employed mean average precision (MAP) \cite{baeza1999modern}, mean reciprocal rank (MRR) \cite{voorhees1999trec}, and precision at position 1 (P@1) as evaluation metrics. We did not calculate R$_2$@1 because in Douban corpus one context could have more than one correct responses, and we have to randomly sample one for R$_2$@1, which may bring bias to evaluation. %We use different evaluation metrics on our Chinese data set, as we believe precision (MAP and P@1) and recall (MRR) are both important in practice. 
	When using the labeled set, we removed conversations with all negative responses or all positive responses, as models make no difference with them. There are $6,670$ context-response pairs left in the test set.

	\begin{table}
		\small
			\scalebox{0.9}{
		\centering
		\begin{tabular}{c|c|c|c}
			\thickhline
			&train&val&test\\ \hline
			$\#$ context-response pairs &1M&50k & 10k\\ \hline
			$\#$  candidates per context & 2&2&10\\ \hline
			$\#$  positive candidates per context  &1&1& 1.18\\ \hline
			Min. $\#$  turns per context  & 3&3&3\\ \hline
			Max. $\# $  turns per context   & 98&91&45\\\hline
			Avg. $\#$ turns per context  & 6.69&6.75&6.45\\\hline
			Avg. $\#$  words per utterance &18.56&18.50 & 20.74\\
			\thickhline
		\end{tabular}}			\vspace{-2mm}
		\caption{Statistics of Douban Conversation Corpus \label{dataset}}	
		\vspace{-8mm}
	\end{table}
	\subsection{Baseline}
	\begin{table*}[t]
		\small
		\centering
		\begin{tabular}{l|c|c|c|c|c|c|c|c|c|c}
			\thickhline &   \multicolumn{4}{c|}{\textbf{Ubuntu Corpus}}    &        \multicolumn{6}{c}{\textbf{Douban Conversation Corpus}}        \\ \hline
			&  R$_2$@1      &  R$_{10}$@1 &  R$_{10}$@2&  R$_{10}$@5 &MAP&MRR&P@1&  R$_{10}$@1 & R$_{10}$@2  & R$_{10}$@5\\ \hline
			%Random & 0.500 & 0.100 & 0.200 & 0.500 \\
			%Cosine & 0.681 & 0.383 & 0.482 & 0.686\\
			%	Translation model & 0.721 & 0.393 & 0.507 & 0.727\\  
			TF-IDF  & 0.659 & 0.410 & 0.545 & 0.708 & 0.331 &0.359 &0.180 & 0.096&0.172& 0.405 \\ 
			RNN  & 0.768 & 0.403 & 0.547 & 0.819 & 0.390 &0.422 &0.208&0.118&0.223&0.589\\ 
			CNN & 0.848 & 0.549 & 0.684 & 0.896 & 0.417 &0.440 &0.226&0.121&0.252&0.647\\ 
			%LSTM \cite{lowe2015ubuntu} & 0.878 & 0.604 & 0.745 & 0.926 & 0.480 &0.517 &0.310\\ 
			LSTM & 0.901 & 0.638 & 0.784 & 0.949 & 0.485 & 0.527 &0.320&0.187&0.343&0.720\\ 
			BiLSTM & 0.895 & 0.630 & 0.780 & 0.944 &0.479&0.514&0.313&0.184&0.330&0.716\\ \hline
			Multi-View  & 0.908 & 0.662 & 0.801 & 0.951 &0.505&0.543&0.342&0.202&0.350&0.729\\ 
			%	Word-Seq-View\cite{zhou2016multi}  & 0.885&0.608&0.757&0.931 & 0.483 &0.527 &0.328\\
			%	UtterSeq-View\cite{zhou2016multi}  & 0.888 & 0.621 &0.765&0.934&0.486&0.525&0.320\\
			DL2R  & 0.899& 0.626 & 0.783 & 0.944&0.488&0.527&0.330&0.193&0.342&0.705 \\ \hline
			MV-LSTM & 0.906& 0.653 & 0.804 & 0.946& 0.498 & 0.538 & 0.348 &0.202&0.351&0.710 \\ 
			Match-LSTM & 0.904& 0.653 & 0.799 & 0.944& 0.500& 0.537 & 0.345&0.202&0.348&0.720  \\ 
			Attentive-LSTM & 0.903& 0.633 & 0.789 & 0.943& 0.495& 0.523 & 0.331&0.192&0.328&0.718  \\ 
			Multi-Channel & 0.904& 0.656 & 0.809 & 0.942& 0.506 & 0.543 & 0.349 &0.203&0.351&0.709 \\  
			Multi-Channel$_{exp}$& 0.714& 0.368 & 0.497 & 0.745& 0.476 & 0.515 & 0.317 &0.179&0.335&0.691\\ \hline
			%	Sequential Match$_{100d}$ & \textbf{0.924} & 0.711 & 0.838 & \textbf{0.958} &0.526&0.571&0.392\\ 
			SMN$_{last}$ & \textbf{0.923} & \textbf{0.723} & \textbf{0.842} & \textbf{0.956} &\textbf{0.526}&\textbf{0.571}&\textbf{0.393}&\textbf{0.236}&\textbf{0.387}&0.729\\
			SMN$_{static}$ & \textbf{0.927} & \textbf{0.725} & \textbf{0.838} & \textbf{0.962} &\textbf{0.523}&\textbf{0.572}& \textbf{0.387}&\textbf{0.228}&\textbf{0.387}&0.734\\
			SMN$_{dynamic}$ & \textbf{0.926} & \textbf{0.726} & \textbf{0.847} & \textbf{0.961} &\textbf{0.529}&\textbf{0.569}&\textbf{0.397}&\textbf{0.233}&\textbf{0.396}&0.724\\
			\thickhline
		\end{tabular}

		\caption{Evaluation results on the two data sets. Numbers in bold mean that the improvement is statistically significant compared with the best baseline. 
		}		\label{exp:response} \vspace{-5mm}	
	\end{table*}
	We considered the following baselines:
	
	\textbf{Basic models}: models in \cite{lowe2015ubuntu} and \cite{kadlec2015improved} including TF-IDF, RNN, CNN, LSTM and BiLSTM. 
	
	\textbf{Multi-view}: the model proposed by Zhou et al. \shortcite{zhou2016multi} that utilizes a hierarchical recurrent neural network to model utterance relationships. %Utterance view also can be regarded as a variant of Hierarchical Recurrent Neural Network model \cite{serban2016hierarchical}.
	
	\textbf{Deep learning to respond (DL2R)}: the model proposed by Yan et al. \shortcite{DBLP:conf/sigir/YanSW16} that reformulates the message with other utterances in the context.
	
	\textbf{Advanced single-turn matching models}: since  BiLSTM does not represent the state-of-the-art matching model, we concatenated the utterances in a context and matched the long text with a response candidate using more powerful models including MV-LSTM \cite{wan2016match} (2D matching), Match-LSTM \cite{wang2015learning}, Attentive-LSTM \cite{tan2015lstm} (two attention based models), and Multi-Channel which is described in Section \ref{multi-channel}. Multi-Channel is a simple version of our model without considering utterance relationships. We also appended the top 5 tf-idf words in context to the input message, and computed the score between the expanded message and a response with Multi-Channel, denoted as Multi-Channel$_{exp}$.
	
	\subsection{Parameter Tuning}
	For baseline models, if their results are available in existing literature (e.g., those on the Ubuntu corpus), we just copied the numbers, otherwise we implemented the models following the settings in the literatures. All models were implemented using Theano  \cite{2016arXiv160502688short}. Word embeddings were initialized by the results of word2vec \cite{mikolov2013distributed} %\footnote{\url{https://code.google.com/archive/p/word2vec/}} 
	which ran on the training data, and the dimensionality of word vectors is $200$. For Multi-Channel and layer one of our model, we set the dimensionality of the hidden states of GRU as $200$. We tuned the window size of convolution and pooling in $\{(2,2),(3,3)(4,4)\}$ and chose  $(3,3)$ finally. The number of feature maps is $8$.  In layer two, we set the dimensionality of matching vectors and the hidden states of GRU as $50$. The parameters were updated by stochastic gradient descent with Adam algorithm \cite{kingma2014adam} on a single Tesla K80 GPU. The initial learning rate is $0.001$, and the parameters of Adam, $\beta_1$ and $\beta_2$ are $0.9$ and $0.999$ respectively. We employed early-stopping as a regularization strategy. Models were trained in mini-batches with a batch size of $200$, and the maximum utterance length is $50$. We set the maximum context length (i.e., number of utterances) as $10$, because the performance of models does not improve on contexts longer than 10 (details are shown in the Section \ref{analysis}). We padded zeros if the  number of utterances in a context is less than $10$, otherwise we kept the last $10$ utterances. 
	\begin{table*}[h!] \vspace{-2mm} \small
		
		\centering
		\begin{tabular}{l|c|c|c|c|c|c|c|c|c|c}
			\thickhline &   \multicolumn{4}{c|}{\textbf{Ubuntu Corpus}}    &        \multicolumn{6}{c}{\textbf{Douban Conversation Corpus}}        \\ \hline
			&  R$_2$@1      &  R$_{10}$@1 &  R$_{10}$@2&  R$_{10}$@5 &MAP&MRR&P@1 &  R$_{10}$@1 &  R$_{10}$@2&  R$_{10}$@5  \\ \hline
			%Random & 0.500 & 0.100 & 0.200 & 0.500 \\
			%Cosine & 0.681 & 0.383 & 0.482 & 0.686\\
			%	Translation model & 0.721 & 0.393 & 0.507 & 0.727\\  
			Replace$_M$ & 0.905 & 0.661 & 0.799 & 0.950 & 0.503 &0.541 &0.343&0.201&0.364&0.729\\ 
			Replace$_A$ & 0.918 & 0.716 & 0.832 & 0.954 & 0.522&0.565 &0.376&0.220&0.385&0.727\\  \hline
			Only $M_1$ & 0.919 & 0.704 & 0.832 & 0.955 & 0.518 &0.562 &0.370&0.228&0.371&0.737\\ 
			Only $M_2$& 0.921 & 0.715 & 0.836 & 0.956 & 0.521 & 0.565 &0.382&0.232&0.380&0.734\\ \hline				
			SMN$_{last}$ & 0.923 & 0.723 & 0.842 & 0.956 &0.526&0.571&0.393&0.236&0.387&0.729\\ \hline
		\end{tabular}
		\caption{Evaluation results of model ablation. \label{exp:discuss}}	
	\end{table*}
	\begin{figure*}[!h]\vspace{-2mm}
		\centering
		\subfigure[$\mathbf{M}_1$ of $u_1$ and $r$]{
			\label{fig:w1_u1} %% label for first subfigure
			\includegraphics[width=4cm,height=3cm]{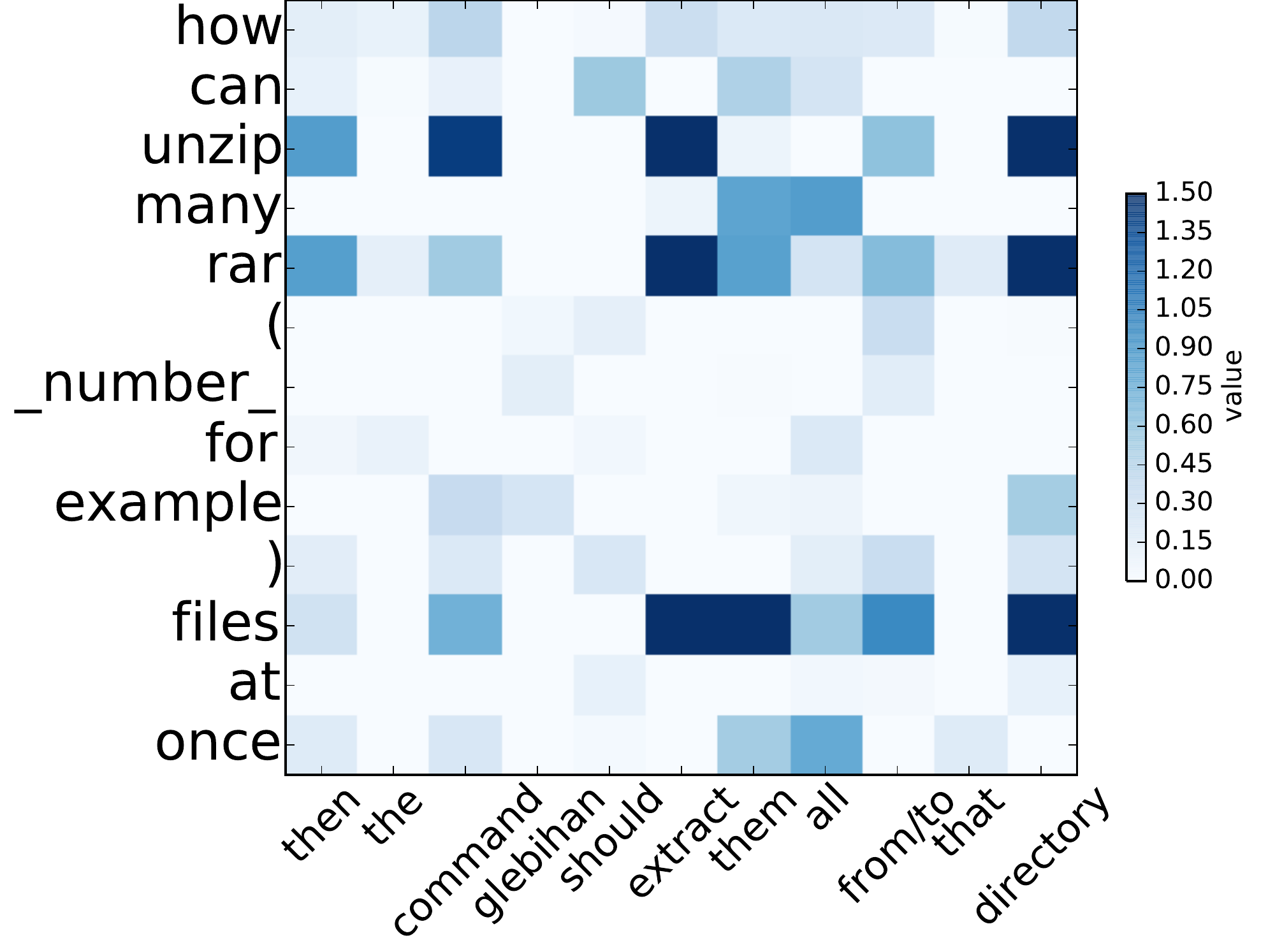}}
		\subfigure[$\mathbf{M}_1$ of $u_3$ and $r$]{
			\label{fig:w1_u3} %% label for second subfigure
			\includegraphics[width=3.8cm,height=3cm]{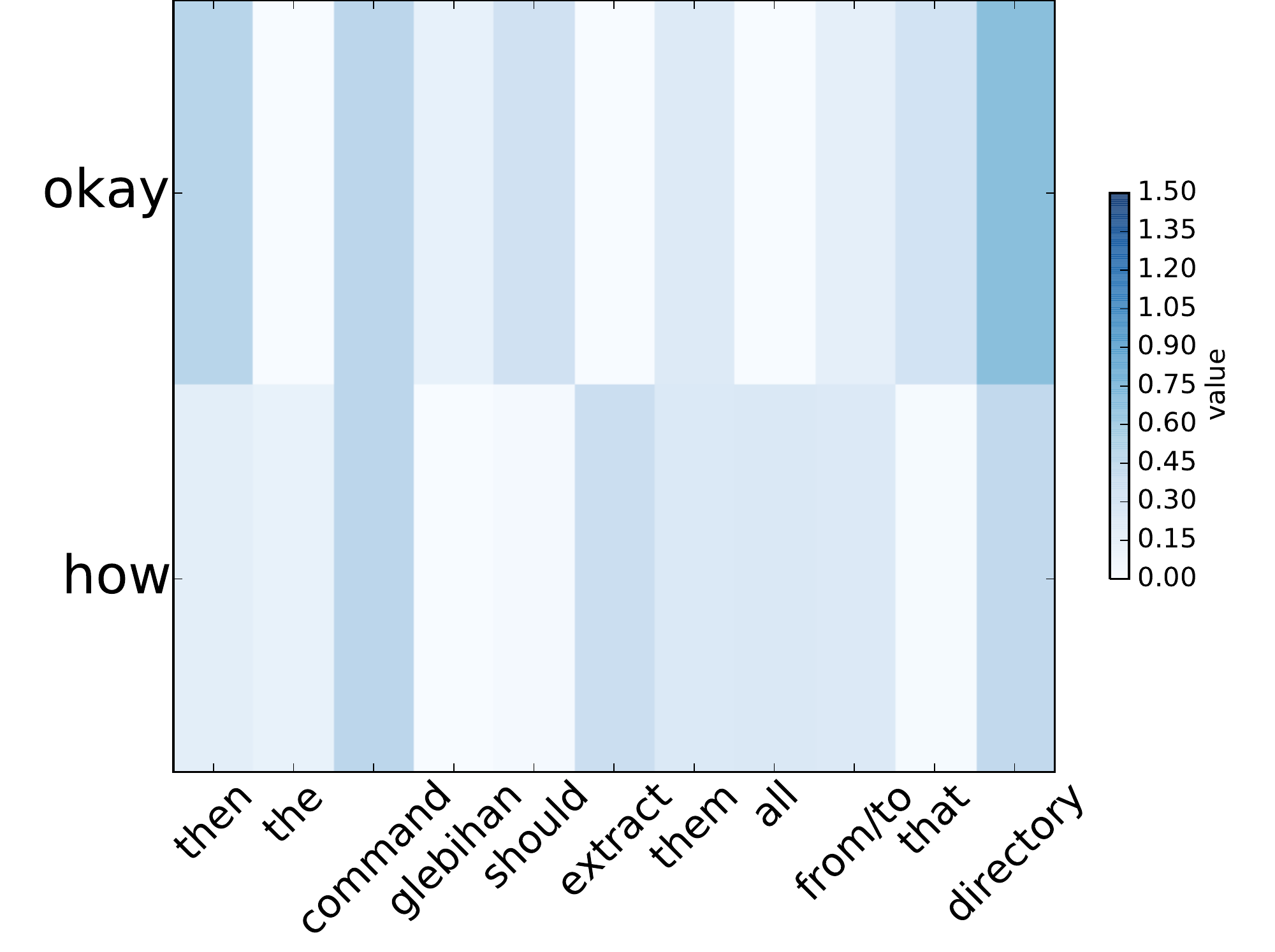}}
		\subfigure[Update gate]{
			\label{fig:update}
			%% label for second subfigure
			\includegraphics[width=4cm,height=3cm]{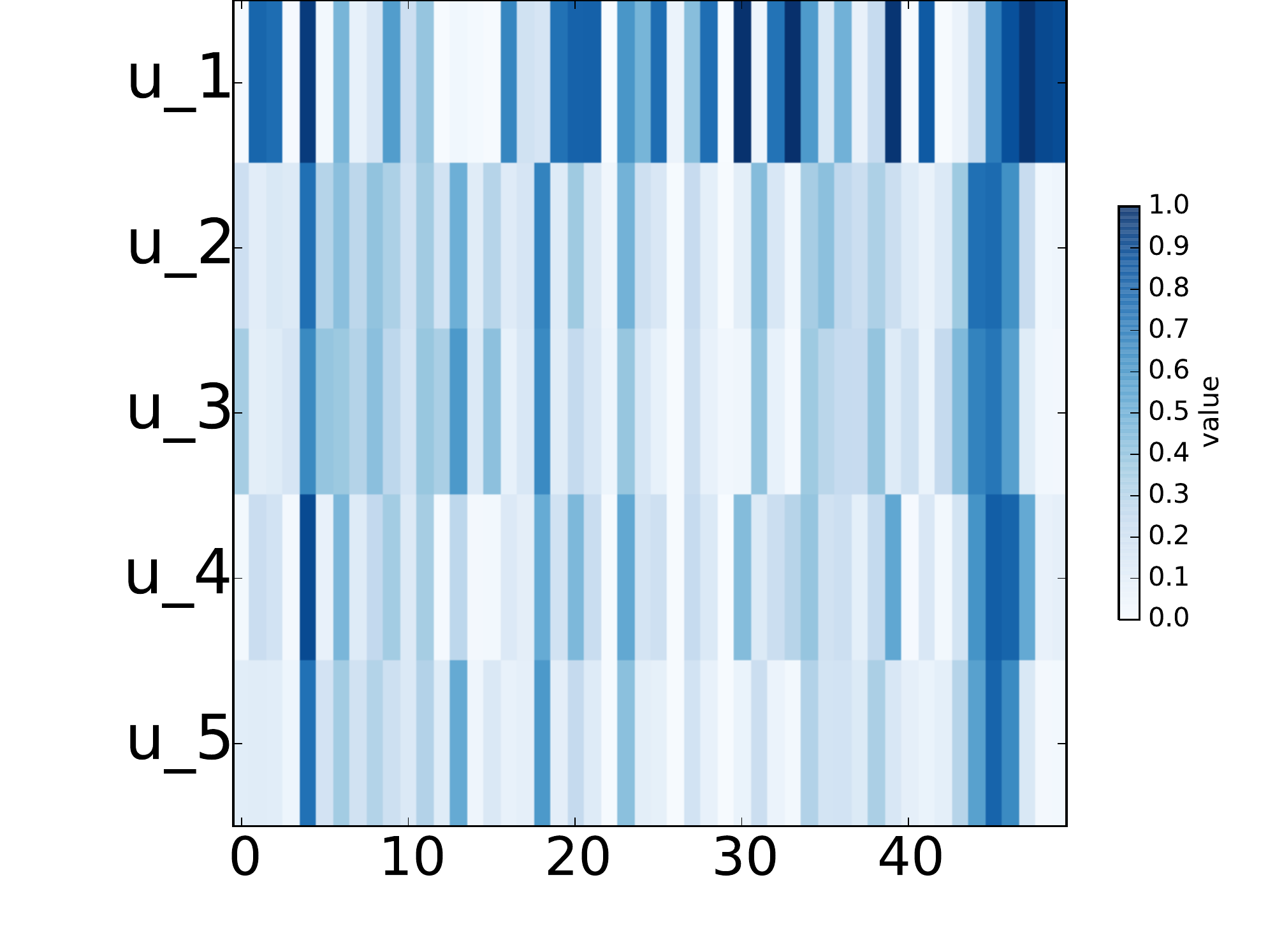}}
		\vspace{-4mm}
		\caption{Model visualization. Darker areas mean larger value.}
		\label{fig:compareall} %% label for entire figure
		\vspace{-4mm}
	\end{figure*}
	\vspace{-2mm}
	
	\subsection{Evaluation Results}
	Table \ref{exp:response} shows the evaluation results on the two data sets.  Our models outperform baselines greatly in terms of all metrics on both data sets, with the improvements being statistically significant (t-test with $p$-value $\leq 0.01$, except $R_{10}@5$ on Douban Corpus). Even the state-of-the-art single-turn matching models perform much worse than our models. The results demonstrate that one cannot neglect utterance relationships and simply perform multi-turn response selection by concatenating utterances together. Our models achieve significant improvements over Multi-View, which justified our ``matching first'' strategy.  DL2R is worse than our models, indicating that utterance reformulation with heuristic rules is not a good method for utilizing context information. $R_n@k$s are low on the Douban Corpus as there are multiple correct candidates for a context (e.g., if there are $3$ correct responses, then the maximum $R_{10}@1$ is $0.33$). 
	SMN$_{dynamic}$ is only slightly better than  SMN$_{static}$ and SMN$_{last}$. The reason might be that the GRU can select useful signals from the matching sequence and accumulate them in the final state with its gate mechanism, thus the efficacy of an attention mechanism is not obvious for the task at hand.
	\vspace{-2mm} 	
	\subsection{Further Analysis}\label{analysis}
	\textbf{Visualization}: we visualize the similarity matrices and the gates of GRU in layer two using an example from the Ubuntu corpus to further clarify how our model identifies important information in the context and how it selects important matching vectors with the gate mechanism of GRU as described in Section \ref{multi-channel} and Section \ref{match_acum}. The example is \emph{ $\{$$u_1$: how can unzip many rar ( $\_number\_$ for example ) files at once; $u_2$: sure you can do that in bash; $u_3$: okay how? $u_4$: are the files all in the same directory? $u_5$: yes they all are; $r$: then the command glebihan should extract them all from/to that directory$\}$}. It is from the test set and our model successfully ranked the correct response to the top position. Due to space limitation, we only visualized $\mathbf{M}_1$, $\mathbf{M}_2$ and  the update gate (i.e. $z$) in Figure \ref{fig:compareall}. %Other pieces of our model are shown in the supplementary material.
	% Due to space limitation, we only visualized $\mathbf{M}_1$ of $u_1$ and $r$ in Figure \ref{fig:w1_u1}, $\mathbf{M}_1$ of $u_3$ and $r$ in Figure \ref{fig:w1_u3}, and the update gate (i.e. $z$) in Figure \ref{fig:update}. They are already enough to support our analysis. Other pieces of our model are visualized in the supplementary material.
	We can see that in $u_1$ important words including ``unzip'', ``rar'', ``files'' are recognized and carried to matching by ``command'', ``extract'', and ``directory'' in $r$,  while $u_3$ is almost useless and thus little information is extracted from it. $u_1$ is crucial to response selection and nearly all information from $u_1$ and $r$ flows to the hidden state of GRU, while other utterances are less informative and the corresponding gates are almost ``closed'' to keep the information from $u_1$ and $r$ until the final state.   
	
%	\textbf{Model ablation}: we investigate the effect of different parts of our model by removing them one by one from SMN$_{last}$. Table \ref{exp:discuss} reports the results. First, replacing the multi-channel ``2D'' matching with a neural tensor network (NTN) \cite{socher2013reasoning} (denoted as Replace$_M$) makes the performance drop dramatically. This is because NTN only matches a pair by an utterance vector and a response vector and loses important information in the pair. Together with the visualization, we can conclude that ``2D'' matching plays a key role in the ``matching first'' strategy as it can capture the important matching information in each pair with minimal loss. Second, the performance slightly drops when replacing the GRU for matching accumulation with a multi-layer perceptron (denoted as Replace$_A$). This indicates that utterance relationships are useful. Finally, we left only one channel in matching and found that $\mathbf{M}_2$ is a little more powerful than $\mathbf{M}_1$ and we can achieve the best results with both of them (except on $R_{10}@5$ on Douban Corpus).  
	\textbf{Model ablation}: we investigate the effect of different parts of SMN by removing them one by one from SMN$_{last}$, shown in Table \ref{exp:discuss}. First, replacing the multi-channel ``2D'' matching with a neural tensor network (NTN) \cite{socher2013reasoning} (denoted as Replace$_M$) makes the performance drop dramatically. This is because NTN only matches a pair by an utterance vector and a response vector and loses important information in the pair. Together with the visualization, we can conclude that ``2D'' matching plays a key role in the ``matching first'' strategy as it captures the important matching information in each pair with minimal loss. Second, the performance drops slightly when replacing the GRU for matching accumulation with a multi-layer perceptron (denoted as Replace$_A$). This indicates that utterance relationships are useful. Finally, we left only one channel in matching and found that $\mathbf{M}_2$ is a little more powerful than $\mathbf{M}_1$ and we achieve the best results with both of them (except on $R_{10}@5$ on the Douban Corpus).  
	
\begin{figure}[!h] \vspace{-2mm} \centering 
			\includegraphics[width=6cm,height=3.2cm]{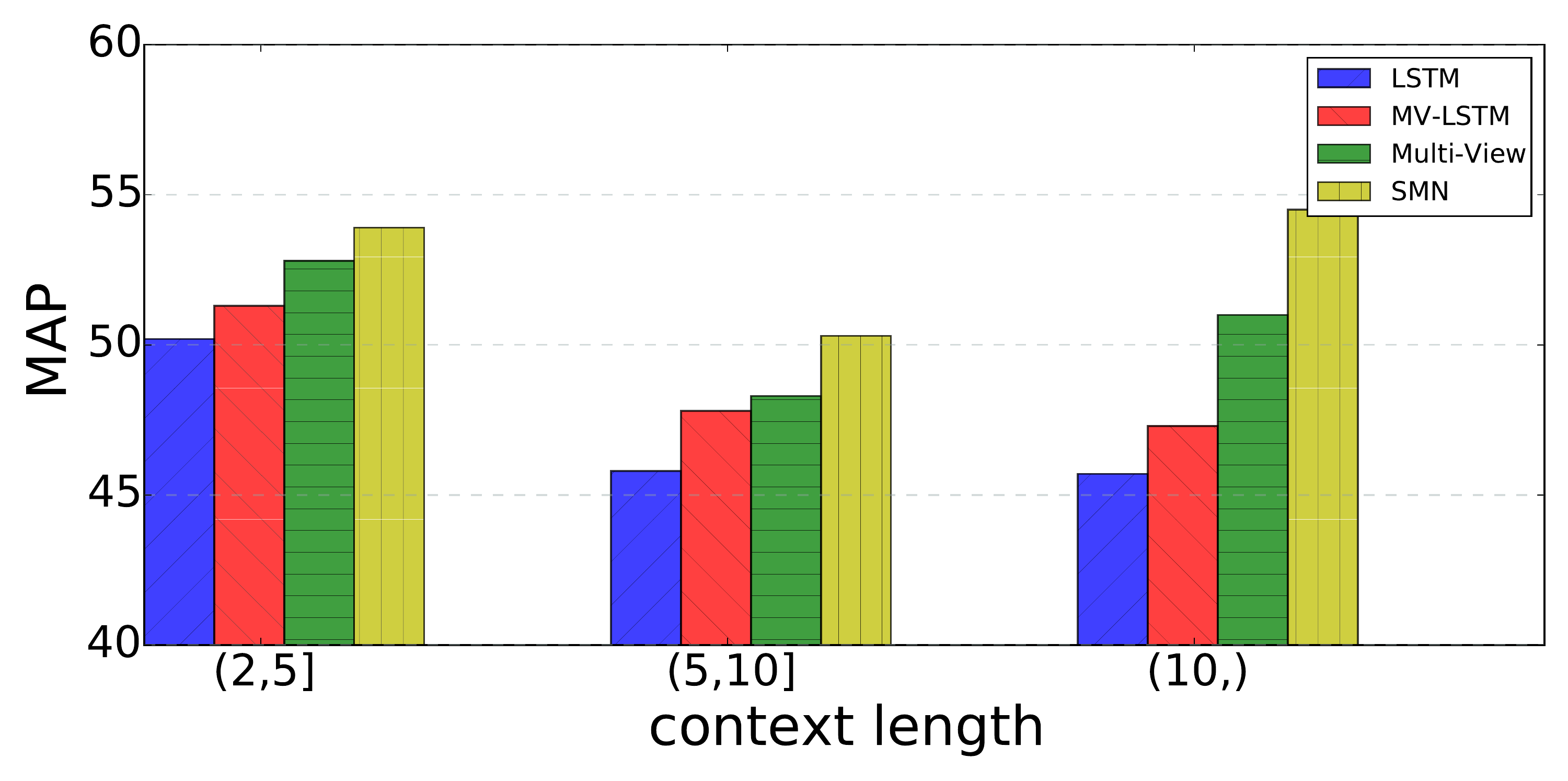}	\vspace{-3mm}			\caption{Comparison across context length}	\label{length} \vspace{-3mm}	
		\end{figure}

	\textbf{Performance across context length}: we study how our model (SMN$_{last}$) performs across the length of contexts. Figure \ref{length} shows the comparison on MAP in different length intervals on the Douban corpus. Our model consistently performs better than the baselines, and when contexts become longer, the gap becomes larger. The results demonstrate that our model can well capture the dependencies, especially long dependencies, among utterances in contexts.% We give the comparisons on other metrics in our supplementary material. 
	
\begin{figure*}[h]
		\centering
		
	\subfigure[Ubuntu Corpus]{
			%% label for first subfigure
			\includegraphics[width=6cm,height=4cm]{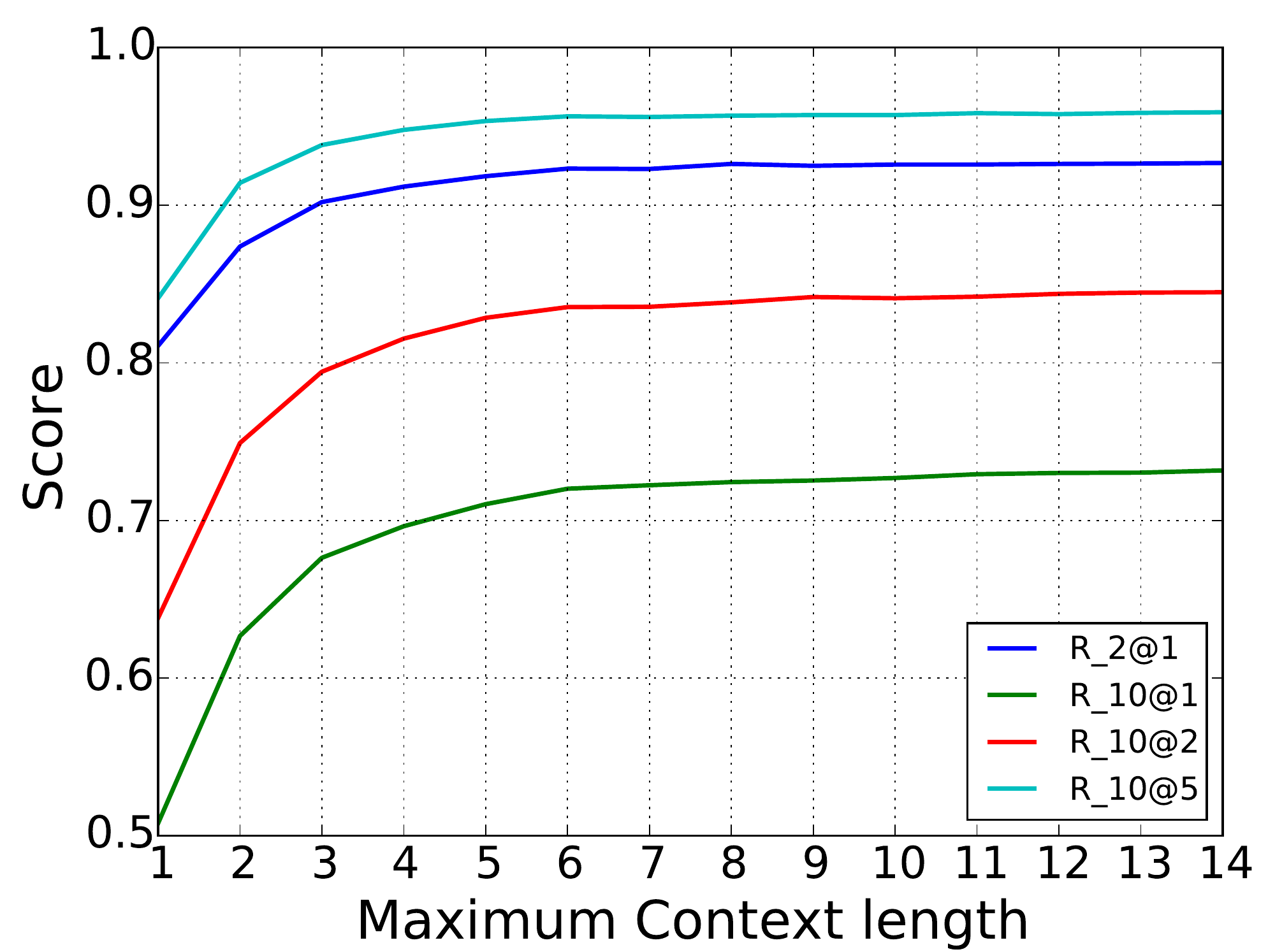}}
		\subfigure[Douban Conversation Corpus]{
			%% label for second subfigure
			\includegraphics[width=6cm,height=4cm]{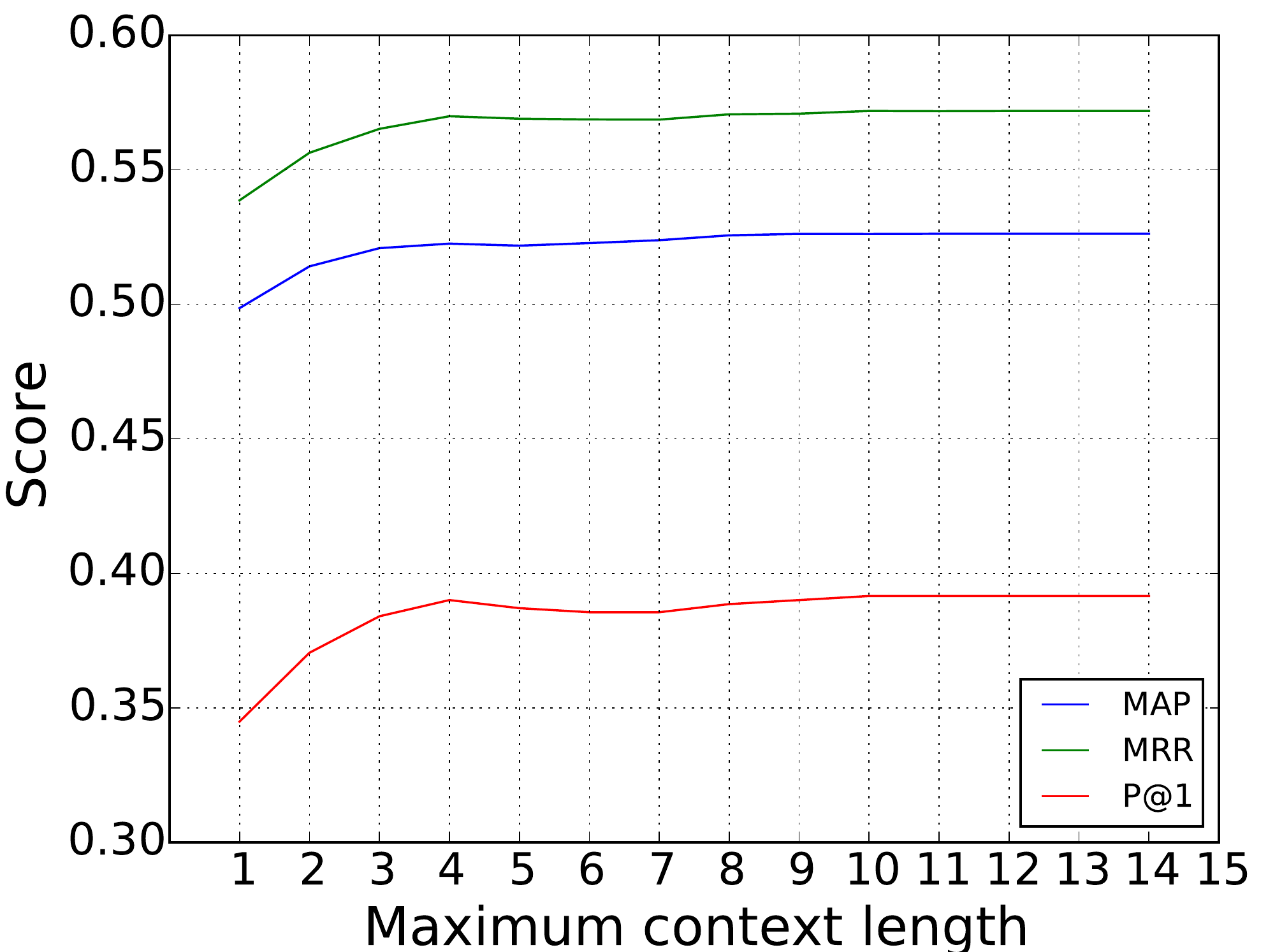}}
		\caption{Performance of SMN across maximum context length}		\label{fig:max_smn_length}
	\end{figure*}

%	\textbf{Retrieval v.s. Generation}: we compared SMN with a state-of-the-art response generation model VHERD \cite{serban2016hierarchical} which was trained using $\mathcal{D}$ on the Douban corpus. We conducted a side-by-side human comparison on the top one responses of the two models for each context in the test set. The result is that SMN wins on $238$ examples, loses on $207$ examples, and is comparable with VHRED on the remaining $555$ examples. This indicates that a retrieval based chatbot with SMN can provide a better experience than the state-of-the-art generation model in practice. 
		\textbf{Maximum context length}: we investigate the influence of maximum context length for SMN. Figure  \ref{fig:max_smn_length} shows the performance of SMN on Ubuntu Corpus and Douban Corpus with respect to maximum context length. From Figure  \ref{fig:max_smn_length}, we find that performance improves significantly when the maximum context length is lower than 5, and becomes stable after the context length reaches 10. This indicates that context information is important for multi-turn response selection, and we can set the maximum context length as 10 to balance effectiveness and efficiency. 

	\textbf{Error analysis}: although SMN outperforms baseline methods on the two data sets, there are still several problems that cannot be handled perfectly.
	
	(1) Logical consistency. SMN models the context and response on the semantic level, but pays little attention to logical consistency. This leads to several DSATs in the Douban Corpus. For example, given a context \emph{\{a: Does anyone know Newton jogging shoes? b: 100 RMB on Taobao. a: I know that. I do not want to buy it because that is a fake which is made in Qingdao ,b: Is it the only reason you do not want to buy it? \}}, SMN gives a large score to the response  \emph{\{ It is not a fake. I just worry about the date of manufacture\}}. The response is inconsistent with the context on logic, as it claims that the jogging shoes are not fake. In the future, we shall explore the logic consistency problem in retrieval-based chatbots.
	
	(2) No correct candidates after retrieval. In the experiment, we prepared 1000 contexts for testing, but only 667 contexts have correct candidates after candidate response retrieval. This indicates that there is still room for candidate retrieval components to improve, and only expanding the input message with several keywords in context may not be a perfect approach for candidate retrieval. In the future, we will consider advanced methods for retrieving candidates.
	
	\section{Conclusion and Future Work}
	We present a new context based model for multi-turn response selection in retrieval-based chatbots. %The architecture matches utterance-response pairs by a multi-channel ``2D'' matching component at first and utilizes utterance relationships to synthesize the pair matching as a session based matching. 
	Experiment results on open data sets show that the model can significantly outperform the state-of-the-art methods. Besides, we publish the first human-labeled multi-turn response selection data set to research communities. In the future, we shall study how to model logical consistency of responses and improve candidate retrieval.

		\section{Acknowledgment}
	We appreciate valuable comments provided by anonymous reviewers and our discussions with Zhao Yan.
	This work was supported by the National Natural Science Foundation of China (Grand Nos. 61672081, U1636211, 61370126), Beijing Advanced Innovation Center for Imaging Technology (No.BAICIT-2016001), National High Technology Research and Development Program of China (No.2015AA016004), and the Fund of the State Key Laboratory of Software Development Environment (No.SKLSDE-2015ZX-16).
	\bibliographystyle{acl_natbib}
	\bibliography{acl2017}
\end{document}